\documentclass[11pt]{article}

\usepackage[in]{fullpage}
\usepackage{times}
\usepackage{graphicx}
\usepackage{cite}
\usepackage[rflt]{floatflt}
\usepackage{epsfig,subfigure,color,multirow}
\usepackage{url}
\usepackage{epstopdf}

\usepackage{enumitem}
\usepackage{amssymb}
\usepackage{graphicx}
\usepackage{multirow}
\usepackage{amsmath}
\usepackage{algorithm}
\usepackage[noend]{algorithmic}
\usepackage{textcomp}
\usepackage{balance}
\usepackage{amsmath}
\usepackage{amsfonts}
\usepackage{float}

\newcommand{\nt}{\noindent}

\newcommand{\bs}[1]{\ensuremath{\boldsymbol{#1}}}
\newcommand{\mc}[1]{\ensuremath{\mathcal{#1}}}
\newcommand{\pth}[1]{\left(#1\right)}
\newcommand{\f}[1]{f\pth{#1}}

\newtheorem{property}{Property}[section]

\begin{document}
\title{\Large Embed and Conquer: \\ Scalable Embeddings for Kernel $k$-Means on MapReduce}

\author{\fontsize{11}{10}\selectfont
Ahmed Elgohary, Ahmed K. Farahat, Mohamed S. Kamel, and Fakhri Karray\\
\fontsize{11}{10}\selectfont University of Waterloo\\
\fontsize{11}{10}\selectfont Waterloo, Canada N2L 3G1 \\
\fontsize{11}{10}\selectfont \{aelgohary, afarahat, mkamel, karray\}@uwaterloo.ca
}

\date{}

\maketitle

\begin{abstract}The kernel $k$-means is an effective method for data clustering which extends the commonly-used $k$-means algorithm to work on a similarity matrix over complex data structures. It is, however, computationally very complex as it requires the complete kernel matrix to be calculated and stored. Further, its kernelized nature hinders the parallelization of its computations on modern scalable infrastructures for distributed computing. In this paper, we are defining a family of kernel-based low-dimensional embeddings that allows for scaling kernel $k$-means on MapReduce via an efficient and unified parallelization strategy. Afterwards, we propose two practical methods for low-dimensional embedding that adhere to our definition of the embeddings family. Exploiting the proposed parallelization strategy, we present two scalable MapReduce algorithms for kernel $k$-means. We demonstrate the effectiveness and efficiency of the proposed algorithms through an empirical evaluation on benchmark datasets.\end{abstract}

\section{\label{Sec:Intro}Introduction}
In today's era of big data, there is an increasing demand from businesses and industries to get an edge over competitors by making the best use of their data. Clustering is one of the powerful tools that data scientists can employ to discover natural groupings from the data. The $k$-means algorithm \cite{Jain99} is the most commonly-used method for tackling this problem. It has gained popularity due to its effectiveness on many datasets as well as the ease of its implementation on different computing architectures.

The $k$-means algorithm, however, assumes that data are available in an attribute-value format, and that all attributes can be turned into numeric values so that each data instance is represented as a vector in some space where the algorithm can be applied. These assumptions are impractical for real data, and they hinder the use of complex data structures in real-world clustering problems. Examples include grouping users in social networks based on their friendship networks, clustering customers based on their behaviour, and grouping proteins based on their structure. Data scientists tend to simplify these complex structures to a vectorized format and accordingly lose the richness of the data they have.

In order to solve these problems, much research has been conducted on clustering algorithms that work on similarity matrices over data instances rather than on a vector representation of the data in a feature space. This led to the advance of different similarity-based methods for data clustering such as the kernel $k$-means \cite{dhillon2007weighted} and the spectral clustering \cite{Luxburg2007}. The focus of this paper is on the kernel k-means. Different from the traditional $k$-means, the kernel $k$-means algorithm works on kernel matrices which encode different aspects of similarity between complex data structures. It has also been shown that the widely-accepted spectral clustering method has an objective function which is equivalent to a weighted variant of the kernel k-means \cite{dhillon2007weighted}, which means that optimizing that criterion allows for an efficient implementation of the spectral clustering algorithm, in which the computationally complex eigendecomposition step is bypassed. Accordingly, the methods proposed in this paper can be leveraged for scaling the spectral clustering method on MapReduce.

The kernel $k$-means algorithm requires calculating and storing the complete
 kernel matrix. Further, all entries of the kernel matrix need to be accessed in each iteration. As a result, the kernel $k$-means suffers from scalability issues when applied to large-scale data. Some recent approaches \cite{Chitta2011, chitta2012efficient} have been proposed to approximate the kernel $k$-means clustering, and allow its application to large data. However, these algorithms are designed for centralized settings, and assume that the data will fit on the memory/disk of a single machine.  

This paper proposes a family of algorithms for scaling the kernel $k$-means over cloud  infrastructures for distributed computing. Such infrastructures tend to be composed of several commodity machines, each of which is of a limited memory and computing power  \cite{viewcloud,Dean2008,mrc}. The machines are connected together in a shared-nothing cluster which means that data transfers between different machines are done through the network. In such settings of infrastructure, ensuring the scalability and fault tolerance of data analysis tasks is troublesome. MapReduce \cite{Dean2008} is a programming model supported by an execution framework that provides scalable and fault tolerant execution of analytical data processing tasks over distributed infrastructures of commodity machines. The proposed algorithms in this paper are designed to perfectly fit into MapReduce programming model and adhere to its computational constraints. We also optimize the execution of the proposed algorithms by considering the different performance aspects of the target computing infrastructure. 

Our approach is based on eliminating the scalability bottlenecks  of the kernel $k$-means by first learning an embedding of the data instances, and then using this embedding to approximate the cluster assignment step in each iteration of the kernel $k$-means algorithm. We show that this approach leads to a unified and MapReduce-efficient scaling strategy. Additionally, we generalize our approach by defining a family of embeddings charactrized by only four properties that ensure the correctness of any embedding method in the defined family for scaling the kernel $k$-means on MapReduce. The contributions of this paper can be summarized as follows.

\begin{itemize}
\item The paper proposes a generic family of embeddings, which we call Approximate Nearest Centroid (APNC) embeddings, and defines its computational and statistical properties that facilitate scaling kernel $k$-means on MapReduce.
\item Explioting the properties of APNC embeddings, we present a unified and efficient parallelization strategy on MapReduce for approximating the kernel $k$-means using any APNC embedding.
\item The paper proposes two instances of APNC embeddings which are based on the Nystr\"om method and the use of $p$-stable distributions for approximating vector norms. 
\item Medium and large-scale experiments have been conducted for comparing the proposed approach to state-of-the-art kernel $k$-means approximations and demonstrating the effectiveness of the parallel algorithms.
\end{itemize}

The paper is organized as follows. Section \ref{Sec:Notations} describes the notations used throughout the paper. Section \ref{Sec:Background} gives a necessary background on MapReduce and the kernel $k$-means algorithm. We define the family of embeddings in Section \ref{Sec:Embeddings}. Then, we describe the proposed kernel $k$-means approximation along with its parallelization strategy on MapReduce in Section \ref{Sec:Approach}. Sections \ref{Sec:Nystrom} and \ref{Sec:APNC} give the details of the two proposed APNC embeddings. Section \ref{Sec:Related} discusses the related work. The experiments and results are shown in Section \ref{Sec:Exp}. Finally, we conclude the paper in Section \ref{Sec:Conc}.

\section{\label{Sec:Notations}Notations}
The following notations are used throughout the paper unless otherwise indicated. Scalars are denoted by small letters (e.g., $m$, $n$), sets are denoted in script letters (e.g., $\mc{L}$), vectors are denoted by small bold italic letters (e.g., $\bs{\phi}$, $\bs{y}$), and matrices are denoted by capital letters (e.g., $\Phi$, $Y$). In addition, the following notations are used:

\nt For a set $\mc{L}$:

\begin{tabular} {p{2cm}p{15cm}}
 $\mc{L}^{(b)}$ &  the subset of $\mc{L}$ corresponding to the data block $b$.\\
 $|\mc{L}|$ &   the cardinality of the set.\\
\end{tabular}

\nt For a vector $\bs{x} \in \mathbb{R}^{m}$:

\begin{tabular} {p{2cm}p{15cm}}
 $\bs{x}_{i}$ &  $i$-th element of $\bs{x}$.\\
 $\bs{x}^{(i)}$ & the vector $\bs{x}$ correponding to the data instance $i$.\\
  $\bs{x}_{[b]}$ & the vector $\bs{x}$ corresponding to the data block $b$.\\
$\| \bs{x} \|_p$ & the $\ell_p$-norm of $\bs{x}$.\\
\end{tabular}

\nt For a matrix $A \in \mathbb{R}^{m \times n}$:

\begin{tabular} {p{2cm}p{15cm}}
 $A_{ij}\;\;$ & $(i,j)$-th entry of $A$.\\
 $A_{i:}$ &  $i$-th row of $A$.\\
 $A_{:j}$ & $j$-th column of $A$.\\
 $A_{\mc{L}:}$, $A_{:\mc{L}}$ & the sub-matrices of $A$ which consist of the set $\mc{L}$ of rows and columns respectively.\\
 $A^{(b)}$ &  the sub-matrix of $A$ corresponding to the data block $b$.\\
\end{tabular}

\section{\label{Sec:Background}Background}
\subsection{\label{Sec:MapReduce}MapReduce Framework}

MapReduce \cite{Dean2008} is a programming model supported by an execution framework for big data analytics over a distributed environment of commodity machines. To ensure scalable and fault-tolerant execution of the analytical jobs, MapReduce imposes a set of constraints on data access at each machine and communication between different machines. MapReduce is currently considered the typical software infrastructure for many data analytics tasks over large distributed clusters.

In MapReduce, the dataset being processed is viewed as distributed chunks of key-value pairs. A single MapReduce job processes the dataset over two phases. In the first phase, namely the \textit{map phase}, each key-value pair is converted by a user defined \textit{map} function to new intermediate key-value pairs. The intermediate key-value pairs of the entire dataset are grouped by the key, and provided to a \textit{reduce} function in the second phase (the \textit{reduce phase)}. The \textit{reducer} processes a single key and its associated values at a time, and outputs new key-value pairs, which are collectively considered the output of the job. For complex analytical tasks, multiple jobs are chained together or multiple iterations of the same job are carried out over the input dataset \cite{Farahat13}. It is important to note that in addition to the processing time of the \textit{map} and \textit{reduce} functions, a major portion of the job execution time is that taken to move the intermediate key-value pairs across the network. Hence, minimizing the size of the intermediate key-value pairs significantly reduces the overall running time of MapReduce jobs. Further, since the individual machines in cloud computing infrastructures are of very limited memory, a scalable MapReduce-algorithm should ensure that the memory required per machine remains within the bound of commodity memory sizes as the data size increases

The simplicity of MapReduce API together with its scalable and fault-tolerant execution framework distinguished MapReduce and its open-source implementation Hadoop \cite{hadoop} as the most attractive paradigm for data analytics tasks on large-scale cloud computing infrastructures. A significant amount of research have been devoted towards scaling complex data analytics algorithms on MapReduce by developing efficient parallelization strategies, or even by introducing novel approximations that lead to MapReduce-efficient algorithms. Recently, various approaches and approximations have been studied and proposed for scaling popular data mining and machine learning algorithms on MapReduce. Such algorithms spanned text mining \cite{textMR}, graph mining \cite{hadi2008,ashraf}, nonnegative matrix factorization \cite{NMFMapReduce}, feature selection \cite{signhLogistc}, regression \cite{robustRegression}, PageRank \cite{pageRankMr} and most recently column subset selection \cite{Farahat13}.

\subsection{\label{Sec:Kernelkmeans}Kernel $k$-Means}
The $k$-means algorithm \cite{Jain99} is the most widely used algorithm for data clustering. The objective of the algorithm is to group the data points into $k$ clusters such that the Euclidean distances between data points in each cluster and that cluster's centroid are minimized. 
An iterative algorithm, namely Lloyd's algorithm \cite{Lloyd1982}, is usually used for the optimization of this criterion function. In each iteration, the Lloyd's algorithm assigns each data point to the nearest centroid and calculates new centroids based on the current assignment of the data points.

The kernel $k$-means \cite{dhillon2007weighted} is a variant of the $k$-means algorithm in which the distance between a data point and a centroid is calculated in terms of the kernel matrix $K$ which encodes the inner-product between data points in some high-dimensional space. Let $\bs{\phi}^{\pth{i}}$ be the representation of a data instance $i$ in the high-dimensional space endowed implicitly by the kernel function $\kappa(.,.)$. In Lloyd's iterations, cluster assignments are made based on the $\ell_2$-distance between $\bs{\phi}^{\pth{i}}$ and each cluster centroid $\bar{\bs{\phi}}^{\pth{c}}$ as
\begin{equation}\label{Eq:Clstr_Assign}
\pi(i) = \arg\min_{c} \left\Vert \bs{\phi}^{\pth{i}}-\bar{\bs{\phi}}^{\pth{c}}\right\Vert_2\:.
\end{equation} Since neither $\bs{\phi}^{(i)}$ nor $\bar{\bs{\phi}}^{(c)}$ can be assumed to be accessible explicitly, the square of the $\ell_2$-distance in Eq. (\ref{Eq:Clstr_Assign}) is expanded in terms of entries from the kernel matrix $K$ as:
\begin{equation}\label{Eq:KKmeans}
\left\Vert
\bs{\phi}^{\pth{i}}-\bar{\bs{\phi}}^{\pth{c}}\right\Vert^2_2
=K_{ii}-\frac{2}{n_{c}}\sum_{a\in\mc{P}_{c}}K_{ia}+\frac{1}{n_{c}^{2}}\sum_{a,b\in\mc{P}_{c}}K_{ab}\:,
\end{equation} where $\mc{P}_{c}$ is the set of instances in cluster $c$, $n_c=\left|\mc{P}_{c}\right|$ and $K_{ab}$ is the ($a$,$b$)-th entry of the kernel matrix.

This means that in order to find the closest centroid for each data instance, a single pass over the whole kernel matrix is needed. In addition, the whole kernel matrix needs to be stored in the memory. This makes the computational and space complexities of the algorithm quadratic. Accordingly, it is infeasible to implement the original kernel $k$-means algorithm on MapReduce due to the limited memory and computing power of each machine. As we increase the data, there will be a scalability bottleneck which limits the application of the kernel $k$-means to large-scale datasets.

\section{\label{Sec:Embeddings}Embeddings for Scaling Kernel $k$-Means}

In this section, we define a family of embeddings, which we call Approximate Nearest Centroid (APNC) embeddings, that can be used to scale kernel $k$-means on MapReduce. Essentially, we aim at embeddings that: (1) can be computed in a MapReduce-efficient manner, and (2) can efficiently approximate the cluster assignment step of the kernel $k$-means on MapReduce (Eq. \ref{Eq:Clstr_Assign}). We start by defining a set of properties which an embedding should have for the aforementioned conditions to be satisfied.

Let $i$ be a data instance, and $\bs{\phi}=\Phi_{:i}$ be a vector corresponding to $i$ in the kernel space implicitly defined by the kernel function. Let $f:\mathbb{R}^{d}\rightarrow\mathbb{R}^{m}$ be an embedding function that maps $\bs{\phi}$ to a target vector $\bs{y}$, i.e., $\bs{y}=f\pth{\bs{\phi}}$.  In order to use $f\left(\bs{\phi}\right)$ with the proposed MapReduce algorithms, the following properties have to be satisfied.
\begin{property}\label{Prp:Linear} $\f{\bs{\phi}}$ is a linear map, i.e., $\bs{y}=\f{\bs{\phi}}=T\bs{\phi}\:,$ where $T\in\mathbb{R}^{m\times d}$.
\end{property}
If this property is satisfied, then for any cluster $c$, the embedding of its centroid is the same as the centroid of the embeddings of the data instances that belong to that cluster:
\[\bs{\bar{y}}^{\pth{c}} =\f{\bs{\bar{\phi}}^{\pth{c}}}=\frac{1}{n_{c}}\sum_{j\in\mc{P}_{c}}\f{\bs{\phi}^{\pth{j}}}=\frac{1}{n_{c}}\sum_{j\in\mc{P}_{c}}\bs{y}^{\pth{j}}\:,\]where $\bar{\bs{y}}^{\pth{c}}$ is the embedding of the centroid $\bs{\bar{\phi}}^{\pth{c}}$.
\begin{property}\label{Prp:Kernel}
$\f{\bs{\phi}}$ is kernelized.
\end{property}
In order for this property to be satisfied, we restrict the columns of the transformation matrix $T$ to be in the subspace of a subset of data instances $\mc{L} \subseteq \mc{D}$, $\left|\mc{L}\right|=l$ and $l \leq n$:
\[ T=R\Phi_{:\mathcal{L}}^{T}\:.\]
Substituting in $\f{\bs{\phi}}$ gives
\begin{equation}\label{Eq:yKernel}
   \bs{y}=\f{\bs{\phi}}=T\bs{\phi}=R\Phi_{:\mathcal{L}}^{T}\bs{\phi}=RK_{\mc{L}i}\:,
\end{equation}where $K_{\mc{L}i}$ is the kernel matrix between the set of instances $\mc{L}$ and the $i$-th data instance, and $R\in\mathbb{R}^{m\times l}$. We refer to $R$ as the embedding coefficients matrix.

Suppose the set $\mathcal{L}$ definied in Property \ref{Prp:Kernel} consists of $q$ disjoint subsets $\mathcal{L}^{(1)}$,$\mathcal{L}^{(2)}$,..., and $\mathcal{L}^{(q)}$. % (i.e. $\mathcal{L} = \bigcup_{b=1}^{q} \mathcal{L}^{(b)}$).
\begin{property}\label{Prp:R} The embedding coefficients matrix $R$ is in a block-diagonal form:  \[R=\left[\begin{array}{ccc}
                R^{\pth{1}} & 0 & 0\\
                      0 & \ddots & 0\\
                      0 & 0 & R^{\pth{q}}
            \end{array}\right]\:,\]where $q$ is the number of blocks and the $b$-th sub-matrix $R^{\pth{b}}$ along with its corresponding subset of data instances $\mathcal{L}^{(b)}$ can be computed and fit in the memory of a single machine.
\end{property}
It should be noted that different embeddings of the defined family differ in their definitions of the coefficients matrix $R$. 

\begin{property}\label{Prp:Error} There exists a function $e\left(\cdot,\cdot\right)$ that approximates the $\ell_2$-distance between each data point $i$ and the centroid of cluster $c$ in terms of their embeddings $\bs{y}^{\pth{i}}$ and $\bs{\bar{y}}^{\pth{c}}$ only, i.e.,
\[
\exists\hspace{0.2cm} e\left(\cdot,\cdot\right):\:\:\left\Vert\bs{\phi}^{\pth{i}}-\bar{\bs{\phi}}^{\pth{c}}\right\Vert_{2}\approx\beta\: e\left(\bs{y}^{\pth{i}},\:\bar{\bs{y}}^{\pth{c}}\right) \:\: \forall \:\: i,c \:\:\:,
\] where $\beta$ is a constant.
\end{property}
This property allows for approximating the cluster assignment step of the kernel $k$-means as
\begin{equation}\label{Eq:NewC}
  \tilde{\pi}(i)=
  \underset{c}{\arg\min}\:\:e\left(\bs{y}^{\pth{i}},\:\bar{\bs{y}}^{\pth{c}}\right)\:.
\end{equation}

\section{\label{Sec:Approach}MapReduce-Efficient Kernel $k$-Means}

\begin{algorithm}[t]
\caption{APNC Embedding on MapReduce \label{Alg:APNCEmbd}}
%\textbf{Input:}$\mathcal{D}$, $\kappa(.,.)$, $R$, $\mathcal{S}$, $q$, %\textbf{Output:} $Y$
\textbf{Input:} Distributed data points $\mathcal{D}$, Kernel function $\kappa(.,.)$, Embedding coefficients matrix $R$, Sample data points $\mathcal{L}$, Number of embedding blocks $q$\\
\textbf{Output:} Embedding matrix $Y$\\
%\textbf{Steps:}
\begin{algorithmic}[1]
	\STATE \textbf{for} $b$ = $1$:$q$	
	\STATE \hspace{0.5cm}\textbf{\textit{map:}}
	\STATE \hspace{1cm}Load $\mathcal{L}^{\left(b\right)}$ and $R^{\left(b\right)}$
	%\STATE \hspace{1cm}$\mathcal{P}$ $\leftarrow$ $\mathcal{L}^{\left(b\right)}$
	\STATE \hspace{1cm}\textbf{foreach} $<i,\mathcal{D}\{i\}>$
	\STATE \hspace{1.5cm}$K_{\mathcal{L}^{\left(b\right)}i}$ $\leftarrow$ $\kappa\left(\mathcal{L}^{\left(b\right)},\mathcal{D}\{i\}\right)$
	\STATE \hspace{1.5cm}$\bs{y}^{\left(i\right)}_{[b]}$ $\leftarrow$ $R^{\left(b\right)} K_{\mathcal{L}^{\left(b\right)}i}$
    \STATE \hspace{1.5cm}emit($i$,$\bs{y}^{\left(i\right)}_{[b]}$)	
	\STATE \hspace{1cm}\textbf{end}
	\STATE \textbf{end}
	\STATE \vspace{0.1cm}\textbf{\textit{map:}}
	\STATE \hspace{0.5cm}\textbf{foreach} $<i$,$\bs{y}^{\left(i\right)}_{[1]}, \bs{y}^{\left(i\right)}_{[2]},...,\bs{y}^{\left(i\right)}_{[q]}>$	
	\STATE \hspace{1cm}$Y_{:i}$ $\leftarrow$ join$\left( \bs{y}^{\left(i\right)}_{[1]}, \bs{y}^{\left(i\right)}_{[2]},...,\bs{y}^{\left(i\right)}_{[q]} \right)$	
	\STATE \hspace{1cm}emit($i$,$Y_{:i}$)
	\STATE \hspace{0.5cm}\textbf{end}    	
\end{algorithmic}
\end{algorithm}

In this section, we show how the four properties of APNC embeddings can be exploited to develop  efficient and unified parallel MapReduce algorithms for kernel $k$-means. We start with the algorithm for computing the corresponding embedding for each data instance, then explain how to use these embeddings for approximating the kernel $k$-means. 

From Property \ref{Prp:Kernel} and Property \ref{Prp:R}, the embedding $\bs{y}^{(i)}$ of a data instance $i$ is given by
\begin{eqnarray}
\bs{y}^{(i)}=\left[\begin{array}{ccc}
                R^{\pth{1}} & 0 & 0\\
                      0 & \ddots & 0\\
                      0 & 0 & R^{\pth{q}}
            \end{array}\right]\:K_{\mathcal{L}i}\:,
\label{y_inR}
\end{eqnarray}for a set of selected data points $\mathcal{L}$. The set $\mathcal{L}$ consists of $q$ disjoint subsets $\mathcal{L}^{(1)}$,$\mathcal{L}^{(2)}$,..., and $\mathcal{L}^{(q)}$. So, the vector $K_{\mathcal{L}i}$ can then be written in the form of $q$ blocks as $K_{\mathcal{L}i} = [K_{\mathcal{L}^{(1)}i}^T K_{\mathcal{L}^{(2)}i}^T \hdots K_{\mathcal{L}^{(q)}i}^T]^T$. Accordingly, the embedding formula in Eq. (\ref{y_inR}) can be written as
\begin{eqnarray}
\bs{y}^{(i)}&=&\left[\begin{array}{ccc}
                R^{\pth{1}} & 0 & 0\\
                      0 & \ddots & 0\\
                      0 & 0 & R^{\pth{q}}
            \end{array}\right]\left[\begin{array}{c}
                K_{\mathcal{L}^{(1)}i}\\
                K_{\mathcal{L}^{(2)}i}\\
                \vdots\\
                K_{\mathcal{L}^{(q)}i}\end{array}\right] 
                = \left[\begin{array}{c}
                R^{\pth{1}}K_{\mathcal{L}^{(1)}i}\\
                R^{\pth{2}}K_{\mathcal{L}^{(2)}i}\\
                \vdots\\
                R^{\pth{q}}K_{\mathcal{L}^{(q)}i}\end{array}\right]\:.
\label{y_inblocks}
\end{eqnarray}

As per Property \ref{Prp:R}, each block $R^{\pth{b}}$ and the sample instances $\mathcal{L}^{(b)}$, used to compute its corresponding $K_{\mathcal{L}^{(b)}i}$, are assumed to fit in the memory of a single machine. This suggests computing $\bs{y}^{(i)}$ in a piecewise fashion,  where each portion ${y}^{(i)}_{[b]}$ is computed separately using its corresponding $R^{\pth{b}}$ and $\mathcal{L}^{(b)}$. 

Our embedding algorithm on MapReduce computes the embedding portions of all data instances in rounds of $q$ iterations. In each iteration, each \textit{mapper} loads the corresponding coefficients block $R^{\pth{b}}$ and data samples $\mathcal{L}^{(b)}$ in its memory. Afterwards, for each data point, the vector $K_{\mathcal{L}^{(b)}i}$ is computed using the provided kernel function, and then used to compute the embedding portion as $\bs{y}^{(i)}_{[b]} = R^{\pth{b}} K_{\mathcal{L}^{(b)}i}$. Finally, in a single \emph{map} phase, the portions of each data instance $i$ are concatenated together to form the embedding $\bs{y}^{(i)}$. It is important to note that the embedding portions of each data point will be stored on the same machine, which means that the concatenation phase has no network cost. The only network cost incurred by the whole embedding algorithm is from loading the sub-matrices $R^{\pth{b}}$ and $\mathcal{L}^{(b)}$ once for each $b$. Algorithm \ref{Alg:APNCEmbd} outlines the embedding steps on MapReduce. We denote each key-value pair of the input dataset $\mc{D}$ as $<\hspace{-0.1cm}i,\mc{D}\{i\}\hspace{-0.07cm}>$ where $i$ refers to the index of the data instance $\mc{D}\{i\}$.

\begin{algorithm} [t]
\caption{APNC Clustering on MapReduce\label{Alg:ClusterStep}}
%\textbf{Input:} $Y$, $m$, $k$, $e(.,.)$ %\textbf{Output:} $\bar{Y}$
\textbf{Input:} Distributed embeddings matrix $Y$, Embedding dimensionality $m$, Number of clusters $k$, Discrepancy function $e(.,.)$\\
\textbf{Output:} Cluster centroids $\bar{Y}$\\
%\textbf{Steps:}
\begin{algorithmic}[1]
	\STATE Generate initial $k$ centroids $\bar{Y}$
	\STATE \textbf{repeat} until convergence
	\STATE \hspace{0.5cm}\textbf{\textit{map:}}
	\STATE \hspace{1cm}Load $\bar{Y}$	
	\STATE \hspace{1cm}Initialize $Z\leftarrow[0]_{m\times k}$ and $\mathbf{g}\leftarrow[0]_{k\times 1}$
	\STATE \hspace{1cm}\textbf{foreach} $<i,Y_{:i}>$
    \STATE \hspace{1.5cm}$\hat{c} = \arg\min_{c} e(Y_{:i},\bar{Y}_{:c})$
    \STATE \hspace{1.5cm}$Z_{:\hat{c}} \leftarrow Z_{:\hat{c}}+Y_{:i}$
    \STATE \hspace{1.5cm}$\mathbf{g}_{\hat{c}} \leftarrow \mathbf{g}_{\hat{c}} +1$
   \STATE \hspace{1cm}\textbf{end}
	\STATE \hspace{1cm}\textbf{for} $c$ = $1$:$k$	
    \STATE \hspace{1.5cm}emit($c$,$<Z_{:c},\mathbf{g}_c>$)
	\STATE \hspace{1cm}\textbf{end}
    \STATE \vspace{0.1cm}\hspace{0.5cm}\textbf{\textit{reduce:}}
	\STATE \hspace{1cm}\textbf{foreach} $<c$, $\mathcal{Z}_c$, $\mathcal{G}_c>$
	\STATE \hspace{1.5cm}$\bar{Y}_{:c} \leftarrow \left(\sum_{Z_{:c} \in \mathcal{Z}_c}Z_{:c}\right)/\left(\sum_{\mathbf{g}_c \in \mathcal{G}} \mathbf{g}_c\right)$
	\STATE \hspace{1.5cm}emit($c$,$\bar{Y}_{:c}$)
	\STATE \hspace{1cm}\textbf{end} 	
	\STATE \textbf{end}
\end{algorithmic}
\end{algorithm}

To parallelize the clustering phase on MapReduce, we make use of Properties \ref{Prp:Linear} and  \ref{Prp:Error}. As mentioned in Section \ref{Sec:Kernelkmeans}, in each kernel $k$-means iteration, a data instance is assigned to its closet cluster centroid given by Eq. (\ref{Eq:Clstr_Assign}). Property \ref{Prp:Error} tells us that each data instance $i$ can be approximately assigned to its closest cluster using only its embedding $\bs{y}^{(i)}$ and the embeddings of the current centroids. Further, Property \ref{Prp:Linear} allows us to compute updated embeddings for cluster centroids, using the embeddings of the data instances assigned to each cluster. 

Let $\bar{Y}$ be a matrix whose columns are the embeddings of the current centroids. Our MapReduce algorithm for the clustering phase parallelizes each kernel $k$-means iteration by loading the current centroids matrix $\bar{Y}$ to the memory of each \emph{mapper}, and uses it to assign a cluster ID to each data point represented by its embedding $\bs{y}^{(i)}$. Afterwards, the embeddings assigned to each cluster are grouped and averaged in a separate \emph{reducer}, to find an updated matrix $\bar{Y}$ to be used in the following iteration. To minimize the network communication cost, we maintain an in-memory matrix $Z$ whose columns are the summation of the embeddings of the data instances assigned to each cluster. We also maintain a vector $\bs{g}$ of the number of data instances in each cluster. We only move $Z$ and $\bs{g}$ of each \textit{mapper} across the network to the \emph{reducers} that compute the updated $\bar{Y}$. Algorithm \ref{Alg:ClusterStep} outlines the clustering steps on MapReduce.

\begin{algorithm}[t]
\caption{APNC Coefficients via Nystr\"om Method\label{Alg:NysEmbd}}
\textbf{Input:} Distributed $n$ data instances $\mathcal{D}$, Kernel function $\kappa(.,.)$, Number of samples $l$, Target dimensionality $m$.\\
\textbf{Output:} Sample data instances $\mathcal{L}$, Embedding coefficients matrix $R$.\\
\begin{algorithmic}[1]
	\STATE \textbf{\textit{map:}}
	\STATE \hspace{0.5cm}\textbf{for} $<i$,$\mathcal{D}\{i\}>$
    \STATE \hspace{1cm}with probability $l/n$, emit($0$,$\mathcal{D}\{i\}$)
	\STATE \hspace{0.5cm}\textbf{end}
    \STATE \vspace{0.1cm}\textbf{\textit{reduce:}}
    \STATE \hspace{0.5cm}\textbf{for} $\mathcal{L}$ $\leftarrow$ all values $\mathcal{D}\{i\}$
    \STATE \hspace{1cm}$K_{\mathcal{L}\mathcal{L}} \leftarrow$ $\kappa\left(\mathcal{L},\mathcal{L}\right)$
    \STATE \hspace{1cm}$[\tilde{V},\tilde{\Lambda}] \leftarrow$ eigen($K_{\mathcal{L}\mathcal{L}}$,$m$)
    \STATE \hspace{1cm}$R \leftarrow$ $\tilde{\Lambda}^{-1/2}\tilde{V}^T$
    \STATE \hspace{1cm}emit($<S,R>$)
    \STATE \hspace{0.5cm}\textbf{end}
\end{algorithmic}
\end{algorithm}

\section{\label{Sec:Nystrom}APNC Embedding via Nystr\"om Method}
One way to preserve the objective function of the cluster assignment step given by Eq.   (\ref{Eq:Clstr_Assign}) is to find a low-rank kernel matrix $\tilde{K}$ over the data instances such that $K\approx \tilde{K}$. Using this kernel matrix in Eq. (\ref{Eq:KKmeans}) results in a cluster assignment which is very close to the assignment obtained using the original kernel $k$-means algorithm. If the low-rank approximation $\tilde{K}$ can be decomposed into $W^TW$ where $W\in \mathbb{R}^{m\times n}$ and $m\ll n$, then the columns of $W$ can be directly used as an embedding that approximates the $\ell_2$-distance between data instance $i$ and the centroid of cluster $c$ as
\begin{equation}\label{Eq_DistW}
\left\Vert \bs{\phi}^{\left(i\right)}-\bar{\bs{\phi}}^{\left(c\right)}\right\Vert _{2}\approx\left\Vert \bs{w}^{\left(i\right)}-\bar{\bs{w}}^{\left(c\right)}\right\Vert _{2}\:.
\end{equation}
To prove that, the right-hand side can be simplified to
\begin{eqnarray*}
  \bs{w}^{\left(i\right)T}\bs{w}^{\left(i\right)}-2\bs{w}^{\left(i\right)T}\bar{\bs{w}}^{\left(c\right)}+\bar{\bs{w}}^{\left(c\right)T}\bar{\bs{w}}^{\left(c\right)}\\
  =\tilde{K}_{ii}-\frac{2}{n_{c}}\sum_{a\in\mc{P}_{c}}\tilde{K}_{ia}+\frac{1}{n_{c}^{2}}\sum_{a,b\in\mc{P}_{c}}\tilde{K}_{ab}
\end{eqnarray*}
The right-hand side is an approximation of the distance function of Eq. (\ref{Eq:KKmeans}). There are many low-rank decompositions that can be calculated for the kernel matrix $K$, including the very accurate eigenvalue decomposition. However, the low-rank approximation to be used has to satisfy the properties defined in Section \ref{Sec:Embeddings}, and accordingly can be implemented on MapReduce in an efficient manner.

One well-known method for calculating low-rank approximations of kernel matrices is the Nystr\"om approximation \cite{Williams01usingthe}. The Nystr\"om method approximates a kernel matrix over all data instances using the sub-matrix of the kernel between all data instances and a few set of data instances $\mc{L}$ as
\begin{equation}\label{Eq:NystromCh3} 
\tilde{K} = D^TA^{-1}D\:,
\end{equation}where $\left|\mc{L}\right|=l\ll n$, $A\in\mathbb{R}^{l\times l}$ is the kernel matrix over the data instances in $\mc{L}$, and $D\in\mathbb{R}^{l\times n}$ is the kernel matrix between the data instances in $\mc{L}$ and all data instances. In order to obtain a low-rank decomposition of $\tilde{K}$, the Nystr\"om method calculates the eigendecomposition of the small matrix $A$ as $A\approx U\Lambda U^T$, where $U\in\mathbb{R}^{l\times m}$ is the matrix whose columns are the leading-$m$ eigenvectors of $A$, and $\Lambda\in\mathbb{R}^{m\times m}$ is the matrix whose diagonal elements are the leading $m$ eigenvalues of $A$. This means that a low-rank decomposition can be obtained as $\tilde{K}=W^TW$ where $W=\Lambda^{-1/2}U^TD$. It should be noted that this embedding satisfies Properties \ref{Prp:Linear} and \ref{Prp:Kernel} as $D=\Phi_{:\mc{L}}^T\Phi$, and accordingly
$\bs{y}^{\pth{i}}=W_{:i}=\Lambda^{-1/2}U^T\Phi_{:\mc{L}}^T\bs{\phi}^{\pth{i}}$. Further, Equation  (\ref{Eq_DistW}) tells us that $e\left(\bs{y}^{\pth{i}},\bs{\bar{y}}^{\pth{c}}\right) = \left\Vert\bs{y}^{\pth{i}}-\bs{\bar{y}^{\pth{c}}}\right\Vert_2$ can be used to approximate the $\ell_2$-distance in Eq. (\ref{Eq:KKmeans}), which satisfies Property \ref{Prp:Error} of the APNC family. 

The embedding coefficient matrix $R=\Lambda^{-1/2}U^T$ is a special case of that described in Property \ref{Prp:R}, which consists of one block of size $m \times l$, where $l$ is the number of instances used to calculate the Nystr\"om approximation, and $m$ is the rank of the eigen-decomposition used to compute both $\Lambda$ and $U$. It can be assumed that $R$ is computed and fits in the memory of a single machine, since an accurate Nystr\"om approximation can usually be obtained using a very few samples and $m \leq l$. Algorithm \ref{Alg:NysEmbd} outlines the MapReduce algorithm of computing the coefficients matrix $R$ based on the Nystr\"om approximation. The algorithm uses the \textit{map}  phase to iterate over the input dataset in parallel, to uniformly sample $l$ data instances. The sampled instances are then moved to a single \textit{reducer} that computes $R$ as described above.

The Nsytr\"om embedding can be extended by the use of the ensemble Nystr\"om method \cite{Kumar2009a}. In that case, each block of $R$ will be the coefficients of the Nystr\"om embedding corresponding to the subset of data instances that belong to that instance of the ensemble. The details of that extension are the subject of a future work.

\begin{algorithm}[t]
\caption{APNC Coefficients via Stable Distributions\label{Alg:SDEmbd}}
\textbf{Input:} Distributed $n$ data instances $\mathcal{D}$, Kernel function $\kappa(.,.)$, Number of samples $l$, Target dimensionality $m$, Tuning parameter $t$.\\
\textbf{Output:} Sample data instances $\mathcal{L}$, Embedding coefficients matrix $R$.\\
\begin{algorithmic}[1]
	\STATE \textbf{\textit{map:}}
	\STATE \hspace{0.5cm}\textbf{for} $<i$,$\mathcal{D}\{i\}>$
    \STATE \hspace{1cm}with probability $l/n$, emit($0$,$\mathcal{D}\{i\}$)
	\STATE \hspace{0.5cm}\textbf{end}
    \STATE \vspace{0.1cm}\textbf{\textit{reduce:}}
    \STATE \hspace{0.5cm}\textbf{for} $\mathcal{L}$ $\leftarrow$ all values $\mathcal{D}\{i\}$
    \STATE \hspace{1cm}$K_{\mathcal{L}\mathcal{L}} \leftarrow$ $\kappa\left(\mathcal{L},\mathcal{L}\right)$
    \STATE \hspace{1cm}$H$ $\leftarrow$ $I - \frac{1}{l}ee^T$
    \STATE \hspace{1cm}$[V,\Lambda] \leftarrow$ eigen($HK_{\mathcal{L}\mathcal{L}}H$)
    \STATE \hspace{1cm}$E$ $\leftarrow$ $\Lambda^{-1/2}V^T$
 	\STATE \hspace{1cm}\textbf{for} $r$ = $1$:$m$	
	\STATE \hspace{1.5cm}$\mathcal{T} \leftarrow$ select $t$ unique values from $1$ to $l$
	\STATE \hspace{1.5cm}$R_{r:}=\sum_{v \in \mathcal{T}} E_{v:}$	
    \STATE \hspace{1cm}\textbf{end}
    \STATE \hspace{1cm}emit($<S,R>$)
    \STATE \hspace{0.5cm}\textbf{end}
\end{algorithmic}
\end{algorithm}

\section{\label{Sec:APNC}APNC Embedding via Stable Distributions}

In this section, we develop our second embedding method based on the results of Indyk \cite{Indyk} which showed that the $\ell_p$-norm of a $d$-dimensional vector $\bs{v}$ can be estimated by means of $p$-stable distributions. Given a $d$-dimensional vector $\bs{r}$ whose entries are i.i.d. samples drawn from a $p$-stable distribution over $\mathbb{R}$, the $\ell_p$-norm of $\bs{v}$ is given by
\begin{equation}
||\bs{v}||_p = \alpha \mathbb{E} [ |\sum_{i=1}^d \bs{v}_i\bs{r}_i| ]\:,
\end{equation} for some positive constant $\alpha$. It is known that the standard Gaussian distribution $\mathcal{N}(0,1)$ is $2$-stable \cite{Indyk}, which means that it can be employed to compute the $\ell_2$-norm of Eq. (\ref{Eq:KKmeans}) as
\begin{equation}
\left\Vert\boldsymbol{\phi}-\boldsymbol{\bar{\phi}}\right\Vert_2 =  \alpha \mathbb{E} [ | \sum_{i=1}^d ( \boldsymbol{\phi}_i - \boldsymbol{\bar{\phi}}_i) \bs{r}_i | ]\:,
\label{euc_expectation}
\end{equation} where $d$ is the dimensionality of the space endowed by the used kernel function and the entries $\bs{r}_i \sim \mathcal{N}(0,1)$. The expectation above can be approximated by the sample mean of multiple values for the term $| \sum_{i=1}^d ( \boldsymbol{\phi}_i - \boldsymbol{\bar{\phi}}_i) r_i |$ computed using $m$ different vectors $\bs{r}$, each of which is denoted as $\bs{r}^{(j)}$. Thus, the $\ell_2$-norm in Eq. (\ref{euc_expectation}) can be approximated as
\begin{equation}
\left\Vert\boldsymbol{\phi}-\boldsymbol{\bar{\phi}}\right\Vert_2 \approx  \frac{\alpha}{m} \sum_{j=1}^m | \sum_{i=1}^d  \left(\boldsymbol{\phi}_i \bs{r}^{(j)}_i - \boldsymbol{\bar{\phi}}_i \bs{r}^{(j)}_i \right)|
\label{euc_sum}
\end{equation}
Define two $m$-dimensional embeddings $\bs{y}$ and $\bs{\bar{y}}$ such that $\bs{y}_j = \sum_{i=1}^d \boldsymbol{\phi}_i \bs{r}^{(j)}_i$ and $\bs{\bar{y}}_j = \sum_{i=1}^d\boldsymbol{\bar{\phi}}_i \bs{r}^{(j)}_i$ or equivalently, $\bs{y}_j =  \boldsymbol{\phi}^T \bs{r}^{(j)}$ and $\bs{\bar{y}}_j = \boldsymbol{\bar{\phi}}^T \bs{r}^{(j)}$. Equation (\ref{euc_sum}) can be expressed in terms of $\bs{y}$ and $\bs{\bar{y}}$ as
\begin{equation}
\left\Vert\boldsymbol{\phi}-\boldsymbol{\bar{\phi}}\right\Vert_2 \approx  \frac{\alpha}{m} \sum_{j=1}^m |\bs{y}_j - \bs{\bar{y}}_j| = \frac{\alpha}{m} \left\Vert\bs{y} - \bs{\bar{y}}\right\Vert_1 \:.
\label{euc_l1}
\end{equation}
Since all of $\boldsymbol{\phi}$, $\boldsymbol{\bar{\phi}}$, and $\bs{r}^{(j)}$ are intractable to explicitly work with, our next step is to kernelize the computations of $\bs{y}$ and $\bs{\bar{y}}$. Without loss of generality, let $\mathcal{T}_j = \{\boldsymbol{\hat{\phi}}^{(1)},\boldsymbol{\hat{\phi}}^{(2)},...,\boldsymbol{\hat{\phi}}^{(t)}\}$ be a set of $t$ randomly chosen data instances embedded and centered into the kernel space (i.e. $\boldsymbol{\hat{\phi}}^{(i)} = \boldsymbol{\phi}^{(i)} - \frac{1}{t}\sum_{j=1}^t \boldsymbol{\phi}^{(j)}$). According to the central limit theorem, the vector $\bs{r}^{(j)} = \frac{1}{\sqrt{t}} \sum_{\boldsymbol{\phi} \in \mathcal{T}_j} \boldsymbol{\phi}$ approximately follows a multivariate Gaussian distribution $\mathcal{N} \left(0,\Sigma \right)$, where $\Sigma$ is the covariance matrix of the underlying distribution of all data instances embedded into the kernel space \cite{KLSH-12}. But according to our definition of $\bs{y}$ and $\bs{\bar{y}}$, the individual entries of $\bs{r}^{(j)}$ have to be independent and identically Gaussians. To fulfil that requirement, we make use of the fact that decorrelating the variables of a joint Gaussian distribution is enough to ensure that the individual variables are independent and marginally Gaussians. Using the whitening transform, $\bs{r}^{(j)}$ is redefined as
\begin{equation}
\bs{r}^{(j)} = \frac{1}{\sqrt{t}} \tilde{\Sigma}^{-1/2}\sum_{\bs{\phi} \in \mathcal{T}^{(j)}} \bs{\phi}\:,
\label{r_decorr}
\end{equation} where $\tilde{\Sigma}$ is an approximate covariance matrix estimated using a sample of $l$ data points embedded into the kernel space and centred as well. We denote the set of the $l$ data points as $\mc{L}$.

With $\bs{r}^{(j)}$ defined as in Eq. (\ref{r_decorr}), the computation of $\bs{y}$ and $\bs{\bar{y}}$ can be fully kernelized using similar steps to those in \cite{KLSH-12}. Accordingly, $\bs{y}$ and $\bs{\bar{y}}$ can be computed as follows: let $K_{\mc{L}\mc{L}}$ be the kernel matrix of $\mc{L}$, and define a centering matrix $H=I -\frac{1}{l}\bs{e}\bs{e}^T$ where $I$ is an $l\times l$ identity matrix, and $\bs{e}$ is a vector of all ones. Denote the inverse square root of the centered version of $K_{\mc{L}\mc{L}}$ as $E$.\footnote{The centered version of $K_{\mc{L}\mc{L}}$ is given by $HK_{\mc{L}\mc{L}}H$. Its inverse square root can be computed as $\Lambda^{-1/2}V^T$ where $\Lambda$ is a diagonal matrix of the eigenvalues of $HK_{\mc{L}\mc{L}}H$ and $V$ is the eigenvector matrix of $HK_{\mc{L}\mc{L}}H$.} The embedding of a vector $\boldsymbol{\phi}$ is then given by
\begin{equation}
\bs{y} = f(\boldsymbol{\phi}) = R\Phi_{:\mc{L}}^T\boldsymbol{\phi}\:,
\label{y_embd}
\end{equation} such that for $j = 1$ to $m$, $R_{j:}= \bs{s}^TE$, where $\bs{s}$ is an $l$-dimensional binary vector indexing $t$ randomly chosen values from 1 to $l$ for each $j$.

Now, we show that the embedding function $f$ defined in Eq. (\ref{y_embd}) is an APNC Embedding function. It is clear from Eq. (\ref{y_embd}) that $f$ is a linear map in a kernelized form which satisfies Properties \ref{Prp:Linear} and \ref{Prp:Kernel}. Equation (\ref{euc_l1}) shows that the $\ell_2$-norm of the difference between a data point $\boldsymbol{\phi}$ and a cluster centroid $\boldsymbol{\bar{\phi}}$ can be approximated up to a constant by $e(\bs{y},\bs{\bar{y}}) = \left\Vert\bs{y}-\bs{\bar{y}}\right\Vert_1$ which satisfies Property \ref{Prp:Error} of the APNC family. The coefficients matrix $R$ in Eq. (\ref{y_embd}) is of a single block, which can be assumed to be computable in the memory of a single commodity machine. That assumption is justified by observing that $R$ is computed using a sample of a few data instances that are used to conceptually estimate the covariance matrix of the data distribution. Furthermore, the target dimensionality, denoted as $m$ in Eq. (\ref{y_embd}), determines the sample size used to estimate the expectation in Eq. (\ref{euc_expectation}), which also can be estimated by a small number of samples. We validate the assumptions about $l$ and $m$ in our experiments. This accordingly satisfies Property \ref{Prp:R}. We outline the MapReduce algorithm for computing the coefficients matrix $R$, defined by Eq.(\ref{y_embd}), in Algorithm \ref{Alg:SDEmbd}. Similar to Algorithm \ref{Alg:NysEmbd}, we sample $l$ data instances in the \textit{map} phase, and then $R$ is computed using the sampled data instances in a single \textit{reducer}.

\section{\label{Sec:Related}Related Work}
The quadratic runtime complexity per iteration, in addition to the quadratic space complexity of the kernel $k$-means have limited its applicability to even medium-scale datasets on a single machine. Recent work \cite{Chitta2011, chitta2012efficient} to tackle these scalability limitations has focused only on centralized settings with the assumption that the dataset being clustered fits into the memory/disk of a single machine. In specific, Chitta \textit{et al.} \cite{Chitta2011} suggested restricting the clustering centroids to an at most rank-$l$ subspace of the span of the entire dataset where $l \ll n$. That approximation reduces the runtime complexity per iteration to $\mc{O}(l^2k+nlk)$, and the space complexity to $\mc{O}(nl)$, where $k$ is the number of clusters. However, that approximation is not sufficient for scaling kernel $k$-means on MapReduce, since assigning each data point to the nearest cluster still requires accessing the current cluster assignment of all data points. It was also noticed by the authors that their method is equivalent to applying the original kernel $k$-means algorithm to the rank-$l$ Nystr\"om approximation of the entire kernel matrix \cite{Chitta2011}. That is algorithmically different from our Nystr\"om-based embedding in the sense that we use the concept of the Nystr\"om approximation to learn low-dimensional embedding for all data instances, which allows for clustering the data instances by applying a simple and MapReduce-efficient algorithm to their corresponding embeddings.

Later, Chitta \textit{et al.} \cite{chitta2012efficient} exploited the Random Fourier Features (RFF) approach \cite{rahimi2007random} to propose fast algorithms for approximating the kernel $k$-means. However, these algorithms inherit the limitations of the RFF approach such as being limited to only shift-invariant kernels, and requiring data instances to be in a vectorized form.  Furthermore, the theoretical and empirical results of Yang \textit{et al.} \cite{yang2012nystr} showed that the kernel approximation accuracy of RFF-based methods depends on the properties of the eigenspectrum of the original kernel matrix, and that ensuring acceptable approximation accuracy requires using a large number of Fourier features, which increases the dimensionality of the computed RFF-based embeddings. In our experiments, we empirically show that our kernel $k$-means methods achieve clustering accuracy superior to those achieved using the state-of-the-art approximations presented in \cite{Chitta2011} and \cite{chitta2012efficient}.

Other than the kernel $k$-means, the spectral clustering algorithm \cite{Luxburg2007} is considered a powerful approach to kernel-based clustering. Chen \textit{et al.} \cite{chen2010parallel} presented a distributed implementation of the spectral clustering algorithm using an infrastructure composed of MapReduce, MPI, and SSTable\footnote{http://wiki.apache.org/cassandra/ArchitectureSSTable}. In addition to the limited scalability of MPI, the reported running times are very large. We believe this was mainly due to the very large network overhead resulting from building the kernel matrix using SSTable. Later,  Gao \textit{et al.} \cite{Hefeeda2012} proposed an approximate distributed spectral clustering approach that relied solely on MapReduce. The authors showed that their approach significantly reduced the clustering time compared to that of Chen \textit{et al.}, \cite{chen2010parallel}. However, in the approach of Gao \textit{et al.}  \cite{Hefeeda2012}, the kernel matrix is approximated as a block-diagonal, which enforces inaccurate pre-clustering decisions that could  result in degraded clustering accuracy.

Scaling other algorithms for data clustering on MapReduce was also studied in recent work \cite{fastclustering,multidimensional,disco}. However, those works are limited to co-clustering algorithms \cite{disco}, subspace clustering \cite{multidimensional}, metric $k$-centers, and metric $k$-median with the assumption that all pairwise similarities are pre-computed and provided explicitly \cite{fastclustering}.

\interfootnotelinepenalty=10000
\section{\label{Sec:Exp}Experiments and Results}

We evaluated the two proposed algorithms by conducting experiments on four medium and three big datasets, called \textbf{USPS}, \textbf{PIE}, \textbf{MNIST}, \textbf{RCV1}, \textbf{CovType}, \textbf{ImageNet-50k}, and the full \textbf{ImageNet}. The \textbf{PIE} dataset is a subset of 11,554 face images, in 68 classes, out of CMU PIE \cite{cmuPie}. Both of the \textbf{USPS} and \textbf{MNIST} datasets are handwritten digits in 10 classes, and their sizes are 9,298 and 70,000, respectively \cite{ChangLibsvm}. The \textbf{RCV1} dataset is a subset of 193,844 news documents, in 103 categories, prepared by Chen \textit{et al.} \cite{chen2010parallel} to evaluate their distributed spectral clustering algorithms. The \textbf{CovType} dataset is a subset of 581,012 observations of cartographic variables. Each observation is associated with one of seven possible forest cover types. The \textbf{ImageNet} dataset is a processed version of the original ImageNet dataset \cite{imagenet_cvpr09} prepared by Chitta \textit{et al.} \cite{Chitta2011} to evaluate their approximate kernel $k$-means approach. In the medium-scale experiments, we used a sample of 50,000 images out of the 1,262,102 images of the \textbf{ImageNet} dataset. That sample dataset is denoted as \textbf{ImageNet-50k}. All the datasets have been used in previous work to evaluate large-scale clustering algorithms in general \cite{chen2011large, chen2010parallel} and the kernel $k$-means algorithm in particular \cite{Chitta2011,chitta2012efficient}. The properties of the datasets are summarized in Table \ref{Tab:Datasets}.

\begin{table}
\caption{\label{Tab:Datasets}The properties of the datasets used in the experiments.
}

\begin{center}
{\small }%
\begin{tabular}{|c|c|c|c|c|}
\hline
\textbf{\footnotesize Dataset} & \textbf{\footnotesize Type} & \textbf{\footnotesize \# Instances} & \textbf{\footnotesize \# Features} & \textbf{\footnotesize \# Clusters}\tabularnewline
\hline
\hline
\textbf{\footnotesize USPS} & {\small Digit Images} & {\small 9,298} & {\small 256} & {\small 10}\tabularnewline
\hline
\textbf{\footnotesize PIE} & {\small Face Images} & {\small 11,554} & {\small 4,096} & {\small 68}\tabularnewline
\hline
\textbf{\footnotesize MNIST} & {\small Digit Images} & {\small 70,000} & {\small 784} & {\small 10}\tabularnewline
\hline
\textbf{\footnotesize RCV1} & {\small Documents} & {\small 193,844} & {\small 47,236} & {\small 103}\tabularnewline
\hline
\textbf{\footnotesize CovType} & {\small Multivariate} & {\small 581,012} & {\small 54} & {\small 7}\tabularnewline
\hline
\textbf{\footnotesize ImageNet} & {\small Images} & {\small 1,262,102} & {\small 900} & {\small 164}\tabularnewline
\hline
\end{tabular}{\small \par}
\end{center}
\end{table}

For all experiments, after the clustering is performed, the cluster labels are compared to ground-truth labels and the Normalized Mutual Information (NMI) \cite{strehl2003cluster} between clustering labels and the class labels is calculated. We also report the embedding time and clustering time of the proposed algorithms in the large-scale experiments.

The medium-scale experiments were carried out using MATLAB on a single machine to demonstrate the effectiveness of the proposed algorithms compared to previously proposed kernel $k$-means approximations. We compared our algorithms - APNC via Nystr\"om (\textit{APNC-Nys}) and APNC via Stable Distributions (\textit{APNC-SD}) - to the approximate Kernel $k$-means approach (\textit{Approx KKM}) \cite{Chitta2011} and the two Random Fourier Features (RFF)-based algorithms (\textit{RFF}) and (\textit{SV-RFF}) presented in \cite{chitta2012efficient}. For \textit{APNC-Nys}, \textit{APNC-SD} and \textit{Approx KKM}, we used three different values for the number of samples $l$, while fixing the parameter $t$ in \textit{APNC-SD} to $40\%$ of $l$ and $m$ to $1000$. For a fair comparison, we set the number of fourier features used in \textit{RFF} and \textit{SV-RFF} to $500$ to obtain $1000$-dimensional embeddings as in \textit{APNC-SD}. An RBF kernel was used for both the \textbf{PIE} and \textbf{Imgnet-50k} datasets. The $\sigma$ parameter was estimated using the self-tuning method used in \cite{Chitta2011}. We used a neural kernel $k(x_1,x_2) = tanh(ax_1^Tx_2 + b)$  for the \textbf{USPS} dataset and a polynomial kernel $k(x_1,x_2) = (x_1^Tx_2 + 1)^d$ for the \textbf{MNIST} dataset. Following \cite{Chitta2011}, the parameters $a,b$ and $d$ were set to $0.0045, 0.11$ and $5$, respectively. Table \ref{tab:Results1} summarizes the average and standard deviation of the NMIs achieved in $20$ different runs of each algorithm. Being limited to only shift-invariant kernels, both \textit{RFF} and \textit{SV-RFF} were only used for the datasets \textbf{PIE} and \textbf{ImageNet-50k}. We also report the clustering accuracy achieved using the exact kernel $k$-means algorithm on the datasets \textbf{PIE} and \textbf{USPS}.

\begin{table*}[!t]
\begin{center}
\caption{\label{tab:Results1}The NMIs (\%) of different kernel $k$-means approximations (single-node experiments). In each sub-table, the best performing approximation(s) for each $l$ according to $t$-test (with $95\%$ confidence level) is highlighted in bold.
}
\begin{tabular}{|c||c|c|c|}
\hline
 & $l=50$ & $l=100$ & $l=300$ \tabularnewline
\cline{2-4}
\multirow{1}{*}{Methods} & \multicolumn{3}{c|}{\textbf{PIE - 11K, RBF}} \tabularnewline
\hline
\textbf{RFF} & $5.2 \pm 0.12$ & $5.2 \pm 0.12$ & $5.2 \pm 0.12$\tabularnewline
\hline
\textbf{SV-RFF} & $5.15 \pm 0.11$ & $5.15 \pm 0.11$ & $5.15 \pm 0.11$\tabularnewline
\hline
\textbf{Approx KKM} & $13.99 \pm 0.6$ & $14.66 \pm 1.01$ & $15.95 \pm 0.83$\tabularnewline
\hline
\textbf{APNC-Nys} & $\mathbf{18.52 \pm 00.26}$ & $\mathbf{19.23\pm 00.36}$ & $\mathbf{20.20\pm00.46}$\tabularnewline
\cline{1-4}
\hline
\textbf{APNC-SD} & $\mathbf{18.62 \pm 0.37}$ & $\mathbf{19.5 \pm 0.38}$ & $\mathbf{20.12 \pm 0.35}$\tabularnewline
\hline
\textbf{Exact KKM} &\multicolumn{3}{c|}{$20.7915\pm 0.4542$}\tabularnewline
\cline{1-4}
\multirow{2}{*} & \multicolumn{3}{c|}{\textbf{ImageNet - 50K, RBF}} \tabularnewline
\hline
\textbf{RFF} & $6.12 \pm 0.04$ & $6.12 \pm 0.04$ & $6.12 \pm 0.04$\tabularnewline
\hline
\textbf{SV-RFF} & $5.96 \pm 0.06$ & $5.96 \pm 0.06$ & $5.96 \pm 0.06$\tabularnewline
\hline
\textbf{Approx KKM} & $14.67 \pm 0.25$ & $15.12 \pm 0.17$ & $15.27 \pm 0.15$\tabularnewline
\hline
\textbf{APNC-Nys} & $\mathbf{15.62\pm 00.17}$ & $\mathbf{15.81\pm 00.12}$ & $\mathbf{15.79\pm00.09}$\tabularnewline
\cline{1-4}
\hline
\textbf{APNC-SD} & $\mathbf{15.66 \pm 0.14}$ & $\mathbf{15.78 \pm 0.14}$ & $\mathbf{15.76 \pm 0.08}$\tabularnewline
\hline
\multirow{2}{*} & \multicolumn{3}{c|}{\textbf{USPS - 9K, Neural}} \tabularnewline
\cline{2-4}
\hline
\textbf{Approx KKM}  & $37.60 \pm 17.50$ & $50.68 \pm 11.28$ & $\mathbf{57.17 \pm 5.44}$\tabularnewline
\hline
\textbf{APNC-Nys} & $\mathbf{51.58\pm 11.74}$ & $\mathbf{55.77\pm 03.30}$ & $\mathbf{58.26\pm 00.95}$\tabularnewline
\cline{1-4}
\hline
\textbf{APNC-SD} & $\mathbf{52.88 \pm 7.25}$ & $\mathbf{55.34 \pm 4.15}$ & $\mathbf{58.22 \pm 0.87}$\tabularnewline
\hline
\textbf{Exact KKM} &\multicolumn{3}{c|}{$59.4367\pm 0.6591$}\tabularnewline
\hline
\multirow{2}{*} & \multicolumn{3}{c|}{\textbf{MNIST - 70K, Polynomial}} \tabularnewline
\cline{2-4}
\hline
\textbf{Approx KKM} & $19.07 \pm 1.45$ & $20.73 \pm 1.30$ & $22.38 \pm 1.06$ \tabularnewline
\hline
\textbf{APNC-Nys} & $19.68\pm 00.71$ & $20.82\pm 01.44$ & $21.93\pm 00.69$ \tabularnewline
\cline{1-4}
\hline
\textbf{APNC-SD} & $\mathbf{23.00 \pm 1.57}$ & $\mathbf{23.08 \pm 1.58}$ & $\mathbf{23.86 \pm 1.82}$ \tabularnewline
\hline
\end{tabular}
\end{center}
\end{table*}

The large-scale experiments carried out using the last three dataset in Table \ref{Tab:Datasets} were conducted on an Amazon EC2\footnote{http://aws.amazon.com/ec2/} cluster which consists of 20 machines. Each machine comes with a memory of 7.5 GB and a two-cores processor. All machines were running Debian 6.0.5, Hadoop version 1.0.3 and Java 1.7.0. We combined our embedding algorithms \textit{APNC-Nys} and \textit{APNC-SD} with the proposed parallelization strategy and compared them to a baseline two-stages method \textit{2-Stage} that uses the exact kernel $k$-means clustering results of a sample of $l$ data instance to propagate the labels to all the other data instances \cite{Chitta2011}. The \textit{2-Stage} method is used as a sanity check to evaluate the relative improvement in clustering accuracy of the \textit{APNC-Nys} and \textit{APNC-SD}.\footnote{The \textit{2-Stage} method was implemented using MATLAB on a single machine since we are only interested in its clustering accuracy.} We evaluated the three algorithms using three different values for $l$ while fixing $m$ in \textit{APNC-Nys} and \textit{APNC-SD} to $500$. We used a self-tuned RBF kernel for all datasets. For simplicity we used a fixed number of 20 iterations in the clustering step as a convergence criteria in both \textit{APNC-Nys} and \textit{APNC-SD}. Table \ref{tab:Results2} summarizes the average and standard deviation of the NMIs achieved in three different runs of each algorithm. The table also reports the embedding and clustering times of the different APNC methods. For each dataset, the clustering time depends only on the dimensionality of the embeddings ($m$). That is why we are reporting a single entry for the clustering time in each dataset.

It can be observed from Table \ref{tab:Results1} that the centralized versions of the proposed algorithms were significantly superior to all the other kernel $k$-means approximations in terms of the clustering accuracy. Both methods performed similarly in all datasets except for the MNIST, in which \textit{APNC-SD} outperformed \textit{APNC-Nys}.  The poor performance of \textit{RFF} and \textit{SV-RFF} is consistent with the results of \cite{yang2012nystr} that showed that for a fixed number of fourier features, the approximation accuracy of RFF-based methods are determined by the properties of the eigenspectrum of the kernel matrix being approximated. The table also shows that when using only 300 samples (i.e. $l = 300$),  \textit{APNC-Nys} and \textit{APNC-SD} achieve very close clustering accuracy to that of the exact kernel $k$-means which confirms the accuracy and the reliability of the proposed approximations.

Table \ref{tab:Results2} demonstrates also the effectiveness of the proposed algorithms in distributed settings compared to the baseline algorithm. The \textit{APNC-SD} managed to outperform the \textit{APNC-Nys} in NMI only on the \textbf{CovType} dataset. Both methods achieved similar NMIs on the two other datasets. It is also worth noting that, to the best of our knowledge, the best reported NMIs in the literature for the datasets \textbf{RCV1}, \textbf{CovType}, and \textbf{ImageNet} are $28.65\%$ using the spectral clustering \cite{chen2010parallel}, $14\%$ using \textit{RFF} \cite{chitta2012efficient} and $10.4\%$ using \textit{Approx-KKM} of \cite{Chitta2011}, respectively. Our algorithms managed to achieve better NMIs on both \textbf{CovType} and \textbf{ImageNet} and a comparable clustering accuracy on \textbf{RCV1}. Table \ref{tab:Results2} also shows that \textit{APNC-Nys} and \textit{APNC-SD} have comparable embedding times. On the other hand, the clustering step of \textit{APNC-SD} is faster than that of \textit{APNC-Nys}, especially in the datasets with a large number of clusters (\textbf{RCV1} and \textbf{ImageNet}). That advantage of the \textit{APNC-SD} algorithm is from using the $\ell_1$-distance as its discrepancy function, while the \textit{APNC-Nys} uses the $\ell_2$-distance as its discrepancy function. To judge the overall efficiency of our algorithms, we compared the total clustering time on the \textbf{RCV1} dataset to the reported clustering running time of the same dataset on a $20$-node cluster in \cite{chen2010parallel}. We are unaware of any reported results for a distributed kernel $k$-means implementation. We are comparing our running times to the running times of the distributed spectral clustering of \cite{chen2010parallel}, to just get a sense of the efficiency of our algorithms. With $l=1500$, the total clustering time using \textit{APNC-SD} was on average $25$ minutes, while the total clustering time of \textit{APNC-Nys} was $29$ minutes. The reported running time for the same dataset on a $20$-nodes cluster in \cite{chen2010parallel} was $95$ minutes.

\begin{table*}
\caption{\label{tab:Results2}The NMIs and run times of different kernel $k$-means approximations (big datasets). In each NMI sub-table, the best performing method(s) for each $l$ according to $t$-test is highlighted in bold. 
}\vspace{-0.7cm}
\begin{center}
\begin{tabular}{|c||c|c|c||c|c|c||c|}
\hline
\multirow{2}{*}{} & \multicolumn{3}{c||}{\textbf{\small NMI (\%)}} & \multicolumn{3}{c||}{\textbf{\small Embedding Time (mins)}} & \textbf{\small Clustering }\tabularnewline
\cline{2-7}
 & {\small $l=500$} & {\small $l=1000$} & {\small $l=1500$} & {\small $l=500$} & {\small $l=1000$} & {\small $l=1500$} & \textbf{\small Time (mins)}\tabularnewline
\hline
\hline
\textbf{\small Methods} & \multicolumn{7}{c|}{\textbf{\small RCV1 - 200K}}\tabularnewline
\hline
{\small 2-Stage} & {\small 13.33\textpm{}00.53} & {\small 13.56\textpm{}00.53} & {\small 13.56\textpm{}00.06} & \multicolumn{3}{c||}{{\small N/A}} & {\small N/A}\tabularnewline
\hline
{\small APNC-Nys} & {\small \textbf{22.15\textpm{}00.09}} & {\small \textbf{23.77\textpm{}00.60}} & {\small \textbf{23.84\textpm{}00.80}} & {\small 03.1\textpm{}00.1} & {\small 05.9\textpm{}00.2} & {\small 10.9\textpm{}00.5} & 19.6\textpm{}00.1\tabularnewline
\hline
{\small APNC-SD} & {\small \textbf{22.21\textpm{}00.39}} & {\small \textbf{24.34\textpm{}00.26}} & {\small \textbf{23.55\textpm{}00.17}} & {\small 03.0\textpm{}00.2} & {\small 05.9\textpm{}00.2} & {\small 09.9\textpm{}00.4} & 15.6\textpm{}00.4\tabularnewline
\hline
\hline
\textbf{\small Methods} & \multicolumn{7}{c|}{\textbf{\small CovType - 580K}}\tabularnewline
\hline
{\small 2-Stage} & {\small 08.95\textpm{}02.98} & {\small 10.23\textpm{}01.07} & {\small 09.85\textpm{}01.88} & \multicolumn{3}{c||}{{\small N/A}} & {\small N/A}\tabularnewline
\hline
{\small APNC-Nys} & {\small 09.53\textpm{}02.55} & {\small 12.31\textpm{}00.74} & {\small 12.51\textpm{}01.08} & {\small 03.7\textpm{}00.2} & {\small 07.4\textpm{}00.2} & {\small 11.3\textpm{}00.5} & 16.4\textpm{}00.1\tabularnewline
\hline
{\small APNC-SD} & {\small \textbf{15.96\textpm{}01.03}} & {\small \textbf{15.08\textpm{}01.40}} & {\small \textbf{15.56\textpm{}00.18}} & {\small 03.8\textpm{}00.2} & {\small 07.2\textpm{}00.1} & {\small 11.6\textpm{}00.3} & 15.8\textpm{}00.1\tabularnewline
\hline
\hline
\textbf{\small Methods} & \multicolumn{7}{c|}{\textbf{\small ImageNet - 1.26M}}\tabularnewline
\hline
{\small 2-Stage} & {\small 07.51\textpm{}00.42} & {\small 07.58\textpm{}00.21} & {\small 07.71\textpm{}00.20} & \multicolumn{3}{c||}{{\small N/A}} & {\small N/A}\tabularnewline
\hline
{\small APNC-Nys} & {\small \textbf{11.33\textpm{}00.05}} & {\small \textbf{11.26\textpm{}00.11}} & {\small \textbf{11.19\textpm{}00.03}} & {\small 15.6\textpm{}00.4} & {\small 30.3\textpm{}02.4} & {\small 45.2\textpm{}01.9} & 63.8\textpm{}02.8\tabularnewline
\hline
{\small APNC-SD} & {\small \textbf{11.27\textpm{}00.06}} & {\small \textbf{11.26\textpm{}00.04}} & {\small \textbf{11.10\textpm{}00.05}} & {\small 14.9\textpm{}01.4} & {\small 30.9\textpm{}00.9} & {\small 44.6\textpm{}01.5} & 23.7\textpm{}00.3\tabularnewline
\hline
\end{tabular}
\vspace{-0.9cm}
\end{center}
\end{table*}

\section{\label{Sec:Conc}Conclusions}

In this paper, we proposed distributed algorithms for scaling kernel $k$-means on MapReduce. We started by defining a family of low-dimensional embeddings characterized by a set of computational and statistical properties. Based on these properties, we presented a unified parallelization strategy that first computes the corresponding embeddings of all data instances of the given dataset. The obtained embeddings are then clustered in a MapReduce-efficient manner. Based on the Nystr\"om approximation and the properties of the stable distributions, we derived two embedding methods that were shown to adhere to the properties of the defined embedding family. Combining each of the two embedding methods with the proposed parallelization strategy, we demonstrated the effectiveness of the presented algorithms by empirical evaluation on medium and large benchmark datasets.

\section*{Acknowledgment}
We thank Radha Chitta and the authors of \cite{Chitta2011} for sharing their processed ImageNet dataset.

\bibliographystyle{abbrv}

\bibliography{References}

\end{document}